\pdfoutput=1

\documentclass[11pt]{article}
\usepackage[dvipdfmx]{graphicx}
\usepackage[whole]{bxcjkjatype}

\usepackage[]{acl}

\usepackage{times}
\usepackage{latexsym}
\usepackage[T1]{fontenc}

\usepackage[utf8]{inputenc}

\usepackage{microtype}
\usepackage{lingmacros}
\usepackage{amsmath,amssymb}
\usepackage{mathtools}
\usepackage{ascmac}
\usepackage{proof,lscape,caption}
\usepackage{multirow}
\usepackage{color}
\usepackage{stmaryrd}
\usepackage{url}
\usepackage[textsize=scriptsize]{todonotes}
\newcommand{\bhline}[1]{\noalign{\hrule height #1}}

\newcommand{\CenterRow}[2]{
  \dimen0=\ht\strutbox%
  \advance\dimen0\dp\strutbox%
  \multiply\dimen0 by#1%
  \divide\dimen0 by2%
  \advance\dimen0 by-.5\normalbaselineskip%
  \raisebox{-\dimen0}[0pt][0pt]{#2}}

\usepackage{graphicx} 
\usepackage{subcaption}
\usepackage{booktabs}
\newcommand{\ours}{\textsc{Jamp} }
\newcommand{\oursno}{\textsc{Jamp}}

\newcommand{\frageasy}{\textsc{fragment\_easy}}
\newcommand{\fraghard}{\textsc{fragment\_hard}}
\newcommand{\formeasy}{\textsc{format\_easy}}
\newcommand{\formhard}{\textsc{format\_hard}}
\newcommand{\short}{\textsc{short}}
\newcommand{\random}{\textsc{random}}

\title{\oursno: Controlled Japanese Temporal Inference Dataset for\\ Evaluating Generalization Capacity of Language Models}

\author{Tomoki Sugimoto$^1$, Yasumasa Onoe$^2$, Hitomi Yanaka$^1$\\
$^1$The University of Tokyo, $^2$The University of Texas at Austin\\
\texttt{\{sugimoto.tomoki,hyanaka\}@is.s.u-tokyo.ac.jp}\\
\texttt{yasumasa@utexas.edu}}

\begin{document}
\maketitle
\begin{abstract}
Natural Language Inference (NLI) tasks involving temporal inference remain challenging for pre-trained language models (LMs).
Although various datasets have been created for this task, they primarily focus on English and do not address the need for resources in other languages.
It is unclear whether current LMs realize the generalization capacity for temporal inference across languages.
In this paper, we present \oursno, a Japanese NLI benchmark focused on temporal inference.
Our dataset includes a range of temporal inference patterns, which enables us to conduct fine-grained analysis.
To begin the data annotation process, we create diverse inference templates based on the formal semantics test suites.
We then automatically generate diverse NLI examples by using the Japanese case frame dictionary and well-designed templates while controlling the distribution of inference patterns and gold labels.
We evaluate the generalization capacities of monolingual/multilingual LMs by splitting our dataset based on tense fragments (i.e., temporal inference patterns).
Our findings demonstrate that LMs struggle with specific linguistic phenomena, such as habituality, indicating that there is potential for the development of more effective NLI models across languages.
\end{abstract}

\section{Introduction}
\label{sec:introduction}

\begin{figure}[t]
    \centering
    \includegraphics[width=7cm]{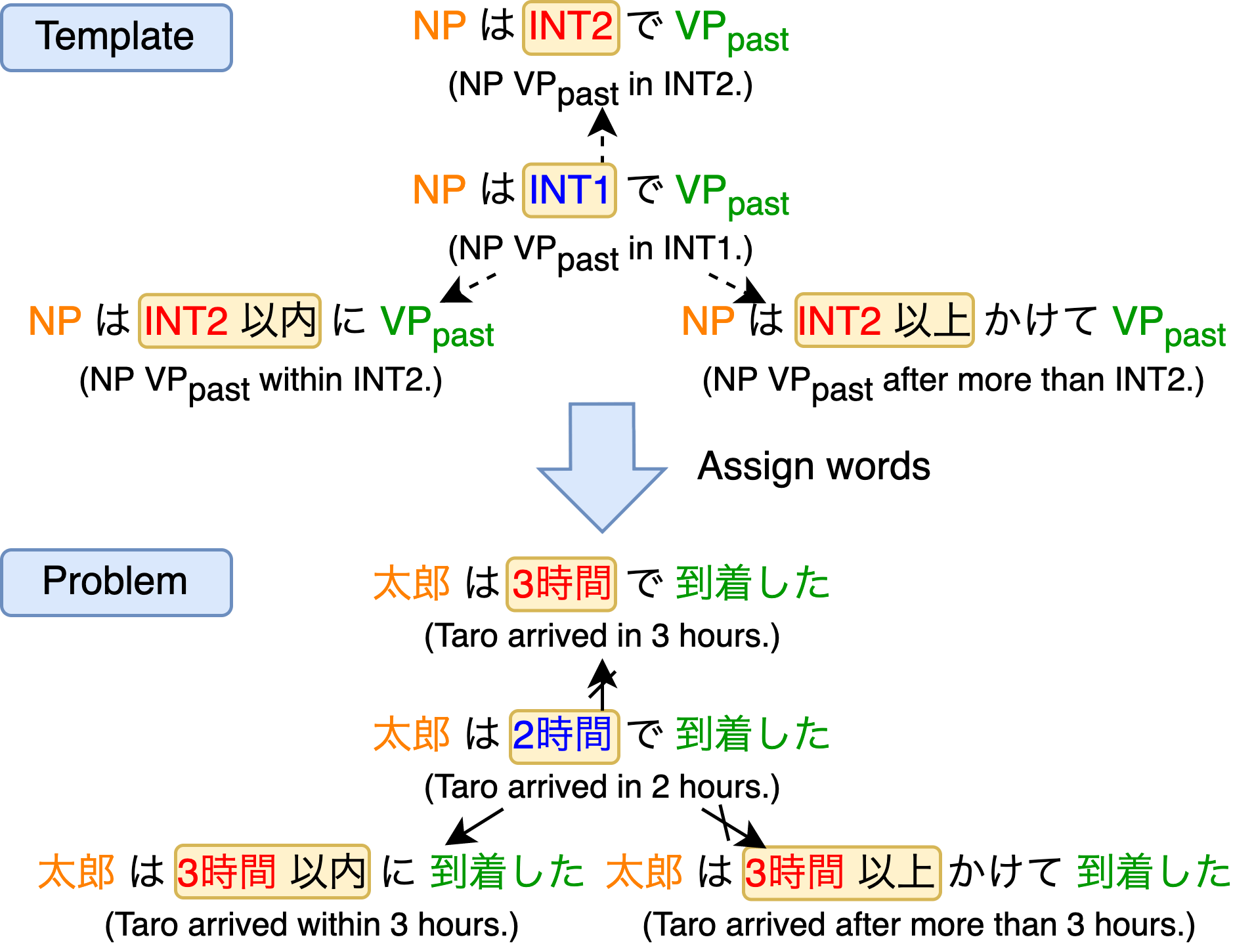}
    \caption{An illustration of our data annotation process. INT in the templates means interval. $\dashrightarrow$ means that the gold label is undetermined, $\rightarrow$ means that the gold label is \textit{Entailment} and $\nrightarrow$ means that the gold label is \textit{Contradiction}.}
    \label{fig:punch}
    \vspace{-20pt}
\end{figure}

\begin{table*}[t]
\centering
\scriptsize
\begin{tabular}{l|l|ll}
\bhline{1.5pt}
Main Tense Fragment &
Sub-tense Fragment &
\multicolumn{2}{l}{Example Problem}\\ \hline\hline
\multirow{5}{*}{Temporal anaphora} & 
\multirow{5}{*}{
\begin{tabular}[c]{@{}l@{}}Reference resolution\\
of 昨日 (\textit{yesterday})
\end{tabular}
} &
P &
\begin{tabular}[c]{@{}l@{}}
\textbf{昨日}\ 、\ APCOM\ は\ 契約書\ に\ 署名した\ 。\\
\textbf{yesterday} , APCOM wa contract ni sign .\\ 
(APCOM signed the contract \textbf{yesterday}.)\\
\textbf{今日}\ は\ 7\ 月\ 14\ 日\ 土曜日\ だ\ 。\\
\textbf{today} wa 7 month 14 day Saturday da .\\
(\textbf{Today} is Saturday, July 14.)
\end{tabular} \\ &                                                              
 & H &
 \begin{tabular}[c]{@{}l@{}}
 APCOM\ は\ 13\ 日\ の\ 金曜日\ に\ 契約書\ に\ 署名した\ 。\\
 APCOM wa 13 day no Friday ni contract ni sign .\\
 (APCOM signed the contract on Friday the 13th.)
 \end{tabular}\\ &                                                                & G &
 Entailment\\ \hline
\multirow{5}{*}{Interval} &
\multirow{5}{*}{Completion of eventuality} &
P &
\begin{tabular}[c]{@{}l@{}}
スミス\ は\ バーミンガム\ に\ \textbf{2\ 年}\ 住んだ\ 。\\
Smith wa Birmingham ni \textbf{2 year} live .\\
(Smith lived in Birmingham \textbf{for two years}.)
\end{tabular}\\  &                                                                 & H &
\begin{tabular}[c]{@{}l@{}}
スミス\ は\ バーミンガム\ に\ 住んだ\ 。\\
Smith wa Birmingham ni live .\\
(Smith lived in Birmingham.)
\end{tabular}\\ &                                                                 & G & Entailment \\
\bhline{1.5pt}
\end{tabular}
\caption{Examples of tense fragments and corresponding problems. P, H, and G indicate a set of premises, a hypothesis, and a gold label, respectively.}
\label{tab:category_examples}
\vspace{-15pt}
\end{table*}

Natural Language Inference (NLI) is the task of determining whether a set of premises entail a hypothesis.
NLI involving temporal inference is a challenging task and remains a significant problem for pre-trained language models (LMs).
One line of research has investigated the temporal inference abilities of LMs~\citep{kober-etal-2019-temporal, vashishtha-etal-2020-temporal, thukral-etal-2021-probing,chen-gao-2022-curriculum}.
However, existing datasets and analyses primarily focus on English, and more analysis and datasets are required for other languages, including Japanese.
Therefore, it is still unclear to what extent current LMs can perform various types of temporal inference across languages.
In this paper, we construct \oursno\footnote{Our dataset is available on \url{https://github.com/tomo-ut/temporalNLI_dataset}}, which is a Japanese NLI dataset for temporal inference, and evaluate the generalization capacity of several LMs on our dataset.

Our goal is to construct a temporal inference dataset that precisely assesses the generalization capacities of LMs.
Manual annotation is a viable option for achieving this goal, but it does not fully meet our needs based on several limitations described below.
Although using crowd-sourcing to increase the size of datasets may be cost-effective~\citep{bowman-etal-2015-large,williams-etal-2018-broad}, managing biases and artifacts in the resulting data can be challenging~\citep{poliak-etal-2018-hypothesis, gururangan-etal-2018-annotation}.
In contrast, datasets manually constructed by experts~\citep{FraCaS,Ai_Kawazoe_2015} may have high quality but are potentially expensive to scale.
Additionally, manual dataset construction makes it difficult to control the distribution of vocabulary and inference patterns in a dataset because it heavily relies on the prior knowledge of each annotator (e.g., word choice).
To address the issues associated with manual annotation, prior work uses template-based approaches that automatically assign diverse vocabulary to templates that are manually created by experts to construct scalable datasets \citep{richardson-etal-2020-probing, yanaka-mineshima-2021-assessing}.
By using this method, we can strictly manage the vocabulary and inference patterns in a dataset, thus it is a suitable approach for probing LMs.

Figure \ref{fig:punch} presents our data annotation process, which consists of two stages: \emph{template creation} and \emph{problem generation}.
We first collect Japanese temporal inference examples from JSeM~\citep{Ai_Kawazoe_2015}, which is the Japanese version of FraCaS~\citep{FraCaS}, and manually transform them into templates by masking content words (e.g., nouns and verbs) and temporal expressions (e.g., date and time), producing 46 tense fragments (i.e., temporal inference patterns) based on formal semantics.
We then generate examples by assigning content words sampled from a Japanese case frame dictionary~\citep{kawahara-kurohashi-2006-fully} and randomly generating temporal expressions to those templates.
These techniques ensure that the sentences in \ours are diverse and cover a wide range of temporal inference patterns.
It is important to note that our temporal NLI examples are derived from a diverse set of templates that are classified with tense fragments, allowing us to create different test splits depending on the goal of evaluation, such as generalization across different tense fragments.

We evaluate two Japanese models and one multilingual model on our dataset.
We analyze whether they can solve our dataset in a zero-shot setting (trained on existing Japanese NLI datasets) and a fine-tuning setting (trained on a small subset of our dataset).
The experimental results demonstrate that the LMs can generalize across different temporal expressions but fail to generalize some tense fragments such as habituality.

\section{Background}
\label{sec:background}
\subsection{Frame}
\label{subsec:frame}

\begin{figure*}[t]
    \centering
    \includegraphics[width=1.0\linewidth]{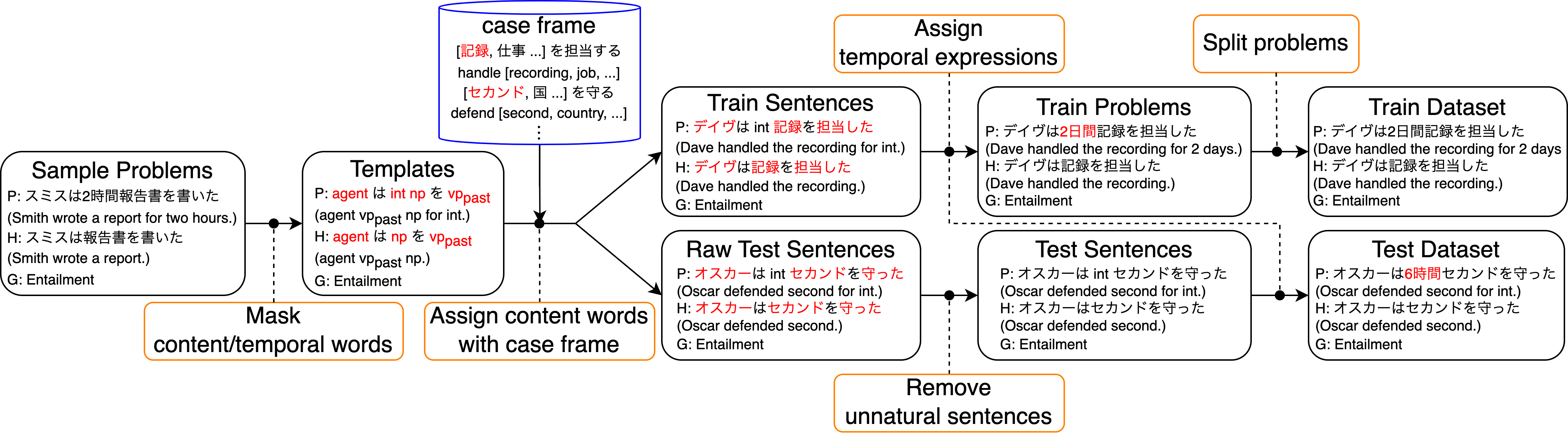}
    \caption{Overview of our data construction pipeline. 1) We first create temporal inference templates from existing examples. 2) We then assign content words using the Japanese case frame dictionary. 3) After isolating train and test examples, we assign temporal expressions to the candidate sentences. Additionally, we manually filter unusable sentences from the test examples.}
    \label{fig:flow}
    \vspace{-15pt}
\end{figure*}

Frame is one of the basic knowledge for language understanding.
There are several English resources for frame knowledge, including VerbNet \citep{schuler-2005-Verbnet}, FrameNet~\citep{baker-etal-1998-berkeley-framenet}, and PropBank \citep{palmer-etal-2005-proposition}, and previous studies have used these resources to construct datasets \citep{poliak-etal-2018-collecting, mitra-etal-2020-enhancing}.

In Japanese, case particles (e.g., が--pronounced \textit{ga}) are attached to verbal arguments (e.g., subject) and determine the case frame.
A Japanese case frame dictionary~\citep{kawahara-kurohashi-2006-fully} is the largest resource that reflects these characteristics of Japanese language.
This case frame dictionary is a set of 110,000 predicates and associated nouns extracted from 10 billion sentences, that are annotated for each predicate usage.
Table~\ref{tab:caseframe} shows an example of a case frame in the Japanese case frame dictionary.

As shown in Table~\ref{tab:caseframe}, the case frame dictionary contains information regarding the frequencies of case frames and nouns.
In this paper, we use these case frames to generate a dataset containing diverse sentence patterns without grammatical errors.

\subsection{Fragments}
\label{subsec:fragments}
Some existing datasets \citep{FraCaS,mccoy-etal-2019-right,yanaka-mineshima-2021-assessing}, including JSeM~\cite{Ai_Kawazoe_2015}, define problem categories for each problem for further analysis.
In this study, we systematically defined tense fragments (i.e., temporal inference patterns) based on the categories of temporal inference patterns in JSeM.

Table~\ref{tab:category_examples} shows some examples of tense fragments (see Appendix~\ref{append:tense_fragment} for additional tense fragments).
In Table~\ref{tab:category_examples}, ``Main Tense Fragment'' represent higher-level classifications, and ``Sub-tense Fragment'' represent sub-classifications that are subdivided from the main tense fragments.
Tense fragments enable a more detailed analysis of LMs' understanding of temporal inference.

\begin{table}[t]
\scriptsize
\centering
\renewcommand{\arraystretch}{1.5}
\begin{tabular}{|ll|}
\hline
\multicolumn{2}{|l|}{\small{到着する (arrive): verb, freq=118520}}             \\ \hline
\multicolumn{1}{|l|}{ga} & 選手 (athlete)$_{freq=205}$, 大統領 (president)$_{freq=114}$, $\cdots$ \\ \hline
\multicolumn{1}{|l|}{ni} & 空港 (airport)$_{freq=24705}$, ホテル (hotel)$_{freq=9639}$, $\cdots$ \\ \hline
\multicolumn{1}{|l|}{$\vdots$} & $\cdots$ \\ \hline
\multicolumn{1}{|l|}{de} & 飛行機 (airplane)$_{freq=347}$, バス (bus)$_{freq=293}$, $\cdots$ \\ \hline
\end{tabular}
\caption{An example of a case frame in the Japanese case frame dictionary.}
\label{tab:caseframe}
\vspace{-20pt}
\end{table}

\section{\oursno}
\label{sec:method}
In this paper, we present \oursno, which is a Japanese NLI dataset for temporal inference, and propose a method for automatic construction from templates based on tense fragments.
Figure \ref{fig:flow} shows the pipeline of our method.
First, we create a template by masking content words and temporal expressions in existing temporal NLI problems (\S\ref{subsec:templatecreation}).
A template consists of the following triplet: (i) a set of premises in which content words and temporal expressions are masked, (ii) a hypothesis in which content words and temporal expressions are masked, and (iii) a condition for determining a gold label.
Here, a gold label can take on three values: \textit{Entailment}, \textit{Contradiction}, and \textit{Neutral}.
Next, we generate training and test sentences by assigning content words selected from the vocabulary list to the template (\S\ref{subsec:problemgeneration}).
We create a vocabulary list by using the Japanese case frame dictionary to make sentences more coherent.\footnote{We considered a generation method using masked LMs or generative models but did not adopt them in this study because the generation time was too long, and it was difficult to control the vocabulary and not change inference patterns and syntactic structures.}

We manually inspect all sentences in the test examples and eliminate any sentences that are unnatural or harmful.
We then generate train and test problems by assigning temporal expressions to train and test sentences.
Finally, we split the training problems along three axes (e.g., tense fragment, time format, and time span) to create training data for various experimental settings (\S\ref{subsec:splitproblem}).
In this section, we describe each of these steps in detail.

\begin{table}[t]
\scriptsize
\centering
\begin{tabular}{l|ll}\bhline{1.5pt}
\multirow{4}{*}{Template} &
P: &
agent\_1 が interval\_1 以内に np\_1 を vp\_1\_past 。\\
&
H: &
agent\_1 は interval\_2 以内に np\_1 を vp\_1\_past 。 \\
&
G: &
\begin{tabular}[c]{@{}l@{}}if interval\_1 $\le$ interval\_2 then Entailment \\else Neutral\end{tabular} \\\hline
\multirow{5}{*}{\begin{tabular}[c]{@{}l@{}}Generated\\Problem\end{tabular}} &
P: &
\begin{tabular}[c]{@{}l@{}}
エレン\ が\ 6\ 年間\ 以内\ に\ ゴール\ を\ 達成した\ 。	\\
Ellen ga 6 years within ni goal o achieved .\\
(Ellen has achieved her goal within six years.)
\end{tabular} \\
&
H: &
\begin{tabular}[c]{@{}l@{}}
エレン\ は\ 5\ 年間\ 以内\ に\ ゴール\ を\ 達成した\ 。\\
Ellen wa 5 years within ni goal o achieved .\\
(Ellen has achieved her goal within five years.)
\end{tabular} \\
&
G: &
Neutral
\\\bhline{1.5pt}
\end{tabular}
\caption{An example of a template and a problem generated by our method.}
\label{tab:problem_generation}
\vspace{-15pt}
\end{table}

\subsection{Template Creation}
\label{subsec:templatecreation}

In the first step, we construct templates consisting of a set of premises, a hypothesis, and a gold label.
We create templates for temporal problems based on problems in the temporal inference section of JSeM by masking content words such as nouns and verbs (e.g., \textit{スミス} (\textit{Smith}), \textit{住んだ} (\textit{lived})), and temporal expressions (e.g., 7\ 月\ 14\ 日 (\textit{July 14}), 2\ 年 (2 \textit{years})).
Additionally, because the gold label depends on the temporal expression in the sentence, we convert the original gold label into a condition in which the gold label is determined by specifying a temporal expression.
Table \ref{tab:problem_generation} shows an example of the template.
In the example in Table \ref{tab:problem_generation}, the condition is ``if interval\_1 $\le$ interval\_2 then \textit{Entailment} else \textit{Neutral}'' and the gold label is determined according to temporal expressions in interval\_1 and interval\_2.

There can be strong correlations between specific words and labels in examples generated from templates based on certain JSeM problems.
Because such correlations could introduce undesired biases into our dataset, we removed these correlations by constructing new challenging templates for some JSeM problems (see Appendix~\ref{appendix:new_template} for examples).

\subsection{Problem Generation}
\label{subsec:problemgeneration}
We generate problems by filling the masks in templates with various nouns, verbs, and temporal expressions and determining the gold label from these temporal expressions.
We use the Japanese case frame dictionary as a vocabulary for selecting verbs and nouns (\S\ref{subsec:frame}).
In this study, we manually filter about 30 offensive words from verbs whose frequency in the dictionary is greater than 1000 and nouns whose frequency in the dictionary is greater than 100 extracted from the case frame dictionary and use filtered words.

We target two types of temporal expressions in this study: time points (e.g., 8\ 月\ 16\ 日\ 7\ 時 (\textit{August} 16, 7:00)) and intervals (e.g., 3\ ヶ月 (3 \textit{months})).
For time points, we use 10 formats combining year/month/day/hour units: Year (Y), Month (M), Day (D), Hour (H), YM, MD, DH, YMD, MDH, and YMDH.
For intervals, we use four formats: Year, Month, Day, and Hour.

We assign content words and temporal expressions to templates as follows.
First, we randomly select a verb with the case in the template from the case frame dictionary.
Next, we randomly select nouns that the selected verb can take as its case in the template.
Here, we select a noun for a subjective case from a manually created list of common first names (e.g., \textit{Alice} and \textit{Bob}).

Then, if a temporal expression exists in the original problem corresponding to the template, we generate a new temporal expression as follows and assign it to templates.
If the original temporal expression is an interval, we generate an interval by concatenating an integer randomly selected from one to nine according to one of the four formats described above.
If the original temporal expression is a time point, we first randomly select a time point within the range of January 1, 2000, at 0:00 to December 31, 2020, at 24:00.
Then, one of the ten formats described above is applied to the selected time point.
For example, if the MD format is applied to 0:00 on January 1, 2010, then the generated temporal expression will be ``January 1.''

Finally, we assign a gold label by evaluating the condition for the gold label in the template.
Table \ref{tab:problem_generation} shows an example of a template and the problem generated from that template.
In Table \ref{tab:problem_generation}, the condition is ``if interval\_1 $\le$ interval\_2 then \textit{Entailment} else \textit{Neutral}.''
Because the generated temporal expressions for interval\_1 and interval\_2 are 6\textit{年間} (\textit{six years}) and 5\textit{年間} (\textit{five years}), respectively, its gold label is \textit{Neutral}.
To ensure that the distribution of gold labels is approximately uniform, we generate the same number of problems from each pair of a template and a gold label.

\begin{figure}[t]
    \centering
    \begin{subfigure}{1.0\linewidth}
    \centering
    \includegraphics[width=0.8\linewidth]{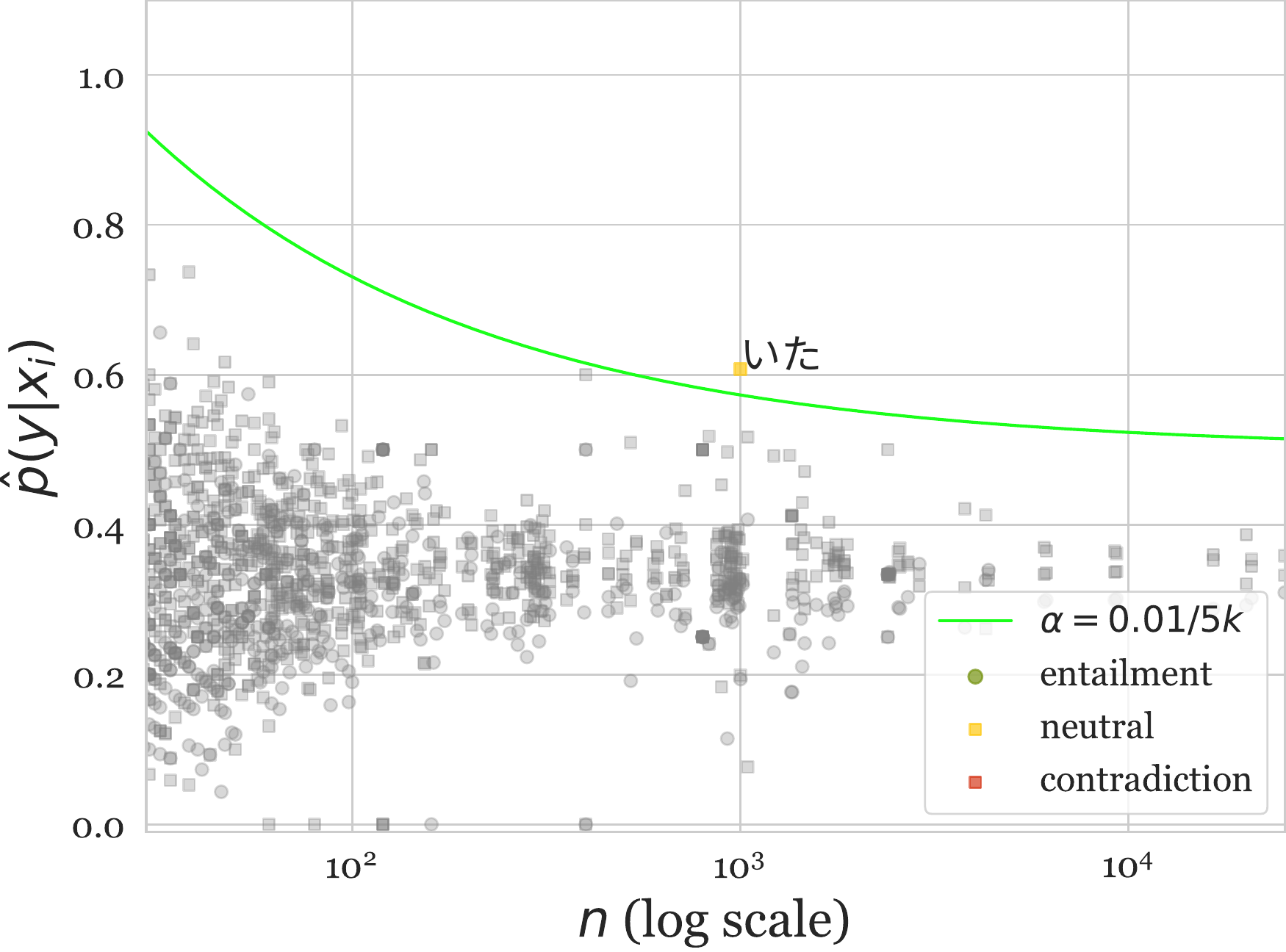}
    \caption{\oursno}
    \end{subfigure} \\
    \vspace{10pt}
    \begin{subfigure}{1.0\linewidth}
    \centering
    \hspace{7pt}
    \includegraphics[width=0.8\linewidth]{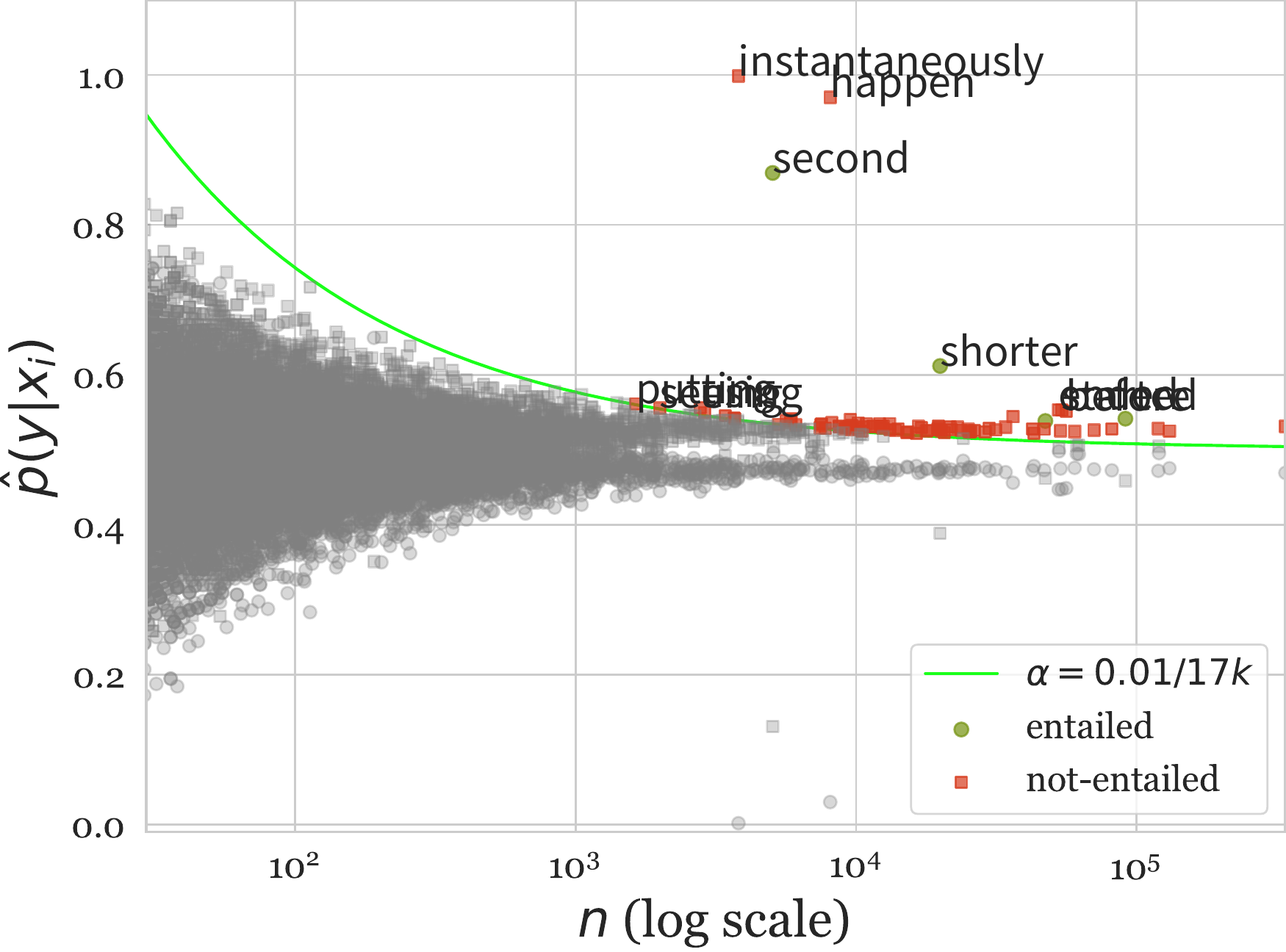}
    \caption{Temporal NLI}
    \end{subfigure}
    \vspace{-10pt}
    \caption{The artifact statistics of (a) \ours and (b) Temporal NLI \citep{vashishtha-etal-2020-temporal} training sets. The majority of words in \oursno, with the exception of ``いた,'' are located below the green line, implying that they do not exhibit spurious correlations with the gold labels. A substantial number of words in Temporal NLI correlate with the gold labels.}
    \label{fig:artifacts}
    \vspace{-15pt}
\end{figure}

\subsection{Quality Control}
\label{subsec:quality_control}

\subsubsection{Dataset Artifacts}
\label{subsubsec:dataset_artifacts}

Previous works have demonstrated that existing datasets are often affected by dataset artifacts and spurious correlations between surface features and gold labels \citep{jia-liang-2017-adversarial, gururangan-etal-2018-annotation, poliak-etal-2018-hypothesis}.
We conduct statistical analysis on our dataset following the method outlined by \citet{gardner-etal-2021-competency} to identify token-level artifacts.
Our analysis reveals the extent to which certain words are highly correlated with one of three labels (see Appendix~\ref{appendix:detail_artifact} for details).

Our automatic data annotation approach enables us to effectively manage the examples that we generate.
We conduct this statistical analysis during the data generation phase and modify vocabulary words and templates to eliminate shortcuts and spurious correlations between certain words and gold labels. 
As depicted in Figure~\ref{fig:artifacts}, the majority of words in \ours do not exhibit spurious correlations with the gold labels, whereas a significant number of words in Temporal NLI \citep{vashishtha-etal-2020-temporal} correlate with the gold labels.\footnote{We sample 100k training examples for this statistical analysis.}
In \oursno, the word ``いた''\footnote{This Japanese word has multiple grammatical roles. One is a past stative verb, and another is a past continuous form of a verb.} stands out as an exception, but its impact is relatively low because its score is close to the green line.

\subsubsection{Dataset Quality}

\paragraph{Naturalness}
We manually check the naturalness of all test examples and filter out disqualified sentences (approx. 40\% of all sentences).\footnote{We ask 3 graduate students studying NLP/linguistics to judge sentence quality.}
Table \ref{tab:unnatural_sentences} shows examples of sentences we remove from the test set and the reasons for their removal.

\begin{table}[t]
\scriptsize
\centering
\begin{tabular}{l|l}\bhline{1.5pt}
Unnatural Sentence & Cause\\\hline\hline
\begin{tabular}[c]{@{}l@{}}
チャーリー\ が\ インク\ を\ 吸った\ 。\\
Charlie ga ink o sucked .\\
(Charlie sucked ink.)
\end{tabular} &
Semantically unnatural\\\hline
\begin{tabular}[c]{@{}l@{}}
ウォルター\ は\ 性格\ に\ 変わった\ 。\\
Walter wa characteristic ni changed .\\
(Walter changed in character.)
\end{tabular} &
Incomplete sentence\\\hline
\begin{tabular}[c]{@{}l@{}}
キャロル\ は\ 速度\ に\ 生ずるていた\ 。\\
Carroll wa speed ni arise .\\
(Carroll arose to speed.)
\end{tabular} &
\begin{tabular}[c]{@{}l@{}}
Semantically unnatural\\
Grammatically unnatural
\end{tabular}
\\\bhline{1.5pt}
\end{tabular}
\caption{Examples of unnatural sentences we filtered.}
\label{tab:unnatural_sentences}
\vspace{-15pt}
\end{table}

Semantically unnatural (e.g., the examples at the top and bottom of Table\ref{tab:unnatural_sentences}) refers to sentences that are grammatically correct but may not be plausible. %
One reason for the generation of such sentences is that the Japanese case frame dictionary does not describe the correspondence between cases (e.g., ヲ格 (accusative) and ニ格 (dative)).
The second case, an incomplete sentence, could be generated since the Japanese case frame dictionary does not describe the essential case for predicates.
Other examples, such as the third, show verbs conjugated in the wrong form.
This is probably because the verb is not included in the dictionary used to conjugate the verb.

\paragraph{Correctness}
We randomly sample 100 cases from the constructed test data and manually judge their entailment labels.
We check whether the judgement is the same as their gold labels.
We confirm that the gold labels in all cases were annotated as intended.
However, the gold labels for some problems were debatable.
For example, in the sentence \textit{I read a book for three hours}, the meaning of \textit{for three hours} can be interpreted as "just three hours," "about three hours," and "at least three hours".
The interpretation depends on the speaker and the context.
In such cases, their gold labels depend on the reading, but we confirmed that they are correct in at least one of the possible readings.

\subsection{Split Problems}
\label{subsec:splitproblem}
Our controlled data generation method enables us to split problems into seen problems (i.e., problems included in both test and training data) and unseen problems (i.e., problems included only in test data) systematically, which is suitable for investigating the generalization capacity of LMs.
In this study, we split our training data to analyze whether LMs can generalize various temporal inference patterns learned from training data.
We split the training data based on three axes: tense fragment, time format, and time span.
Table \ref{tab:seen_unseen_cat}, \ref{tab:seen_unseen_tf}, and \ref{tab:seen_unseen_ts} show an example of a seen/unseen problem in each split.

\begin{table*}[t]
\scriptsize
\centering
\begin{tabular}{l|l|l}\bhline{1.5pt}
         & Seen problem                                                                                                                                                                & Unseen problem                                                                                                                                                   \\\hline\hline
 & TF: Order relation - Transitive, Gold label: Entailment                                                                                                                                     & TF: Order relation - Transitive + Before/After, Gold label: Entailment                                                                                          \\\hline
P        & \begin{tabular}[c]{@{}l@{}}マレット\ は\ イブ\ が\ 出掛ける\ \textbf{前}\ に\ 出掛けた\ 。\\ Mallett wa Eve ga leave \textbf{before} ni leave .\\ (Mullet left \textbf{before} Eve left.)\\イブ\ は\ チャーリー\ が\ 出掛ける\ \textbf{前}\ に\ 出掛けた\ 。\\ Eve wa Charlie ga leave \textbf{before} ni left .\\ (Eve left \textbf{before} Charlie left.)\end{tabular}            & \begin{tabular}[c]{@{}l@{}}マーヴィン\ は\ ペギー\ が\ 留学する\ \textbf{前}\ に\ 留学した\ 。\\ Marvin wa Peggy ga study abroad \textbf{before} ni study abroad .\\ (Marvin studied abroad \textbf{before} Peggy studied abroad.)\\ マーヴィン\ は\ キャロル\ が\ 留学した\ \textbf{後}\ に\ 留学した\ 。\\ Marvin wa Carol ga studied abroad \textbf{after} ni studied abroad .\\ (Marvin studied abroad \textbf{after} Carol studied abroad.)\end{tabular}          \\\hline
H        & \begin{tabular}[c]{@{}l@{}}マレット\ は\ チャーリー\ が\ 出掛ける\ \textbf{前}\ に\ 出掛けた\ 。 \\ Mallett wa Charlie ga leave \textbf{before} ni leave .\\ (Mullet left \textbf{before} Charlie left.)\end{tabular}            & \begin{tabular}[c]{@{}l@{}}ペギー\ は\ キャロル\ が\ 留学した\ \textbf{後}\ に\ 留学した\ 。 \\ Peggy wa Carol ga study abroad \textbf{after} ni study abroad .\\ (Peggy studied abroad \textbf{after} Carol studied abroad.)\end{tabular} \\\hline\hline
 & TF: Usage of 現在 (\textit{now}) - Present tense, Gold label: Entailment                                                                                                                                     & TF: Usage of 現在 (\textit{now}) - Past tense, Gold label: Neutral                                                                                                                             \\\hline
P        & \begin{tabular}[c]{@{}l@{}}マレット\ は\ 皆さん\ に\ 考え方\ を\ 述べ\textbf{ている}\ 。\\ Mallett wa everyone ni thinking o state .\\ (Mallett \textbf{is} stating his thinking to everyone.)\end{tabular}            & \begin{tabular}[c]{@{}l@{}}アイザック\ は\ 見学\ に\ バー\ を\ 訪れ\textbf{ていた}\ 。	 \\ Isaac wa tour ni bar o visit .\\ (Isaac \textbf{was} visiting the bar for a tour.)\end{tabular}          \\\hline
H        & \begin{tabular}[c]{@{}l@{}}マレット\ は\ \textbf{現在}\ 皆さん\ に\ 考え方\ を\ 述べている\ 。\\ Mallett wa \textbf{now} everyone ni thinking o state .\\ (Mallett is \textbf{now} stating his thinking to everyone.)\end{tabular} & \begin{tabular}[c]{@{}l@{}}アイザック\ は\ \textbf{現在}\ 見学\ に\ バー\ を\ 訪れている\ 。\\ Isaac wa \textbf{now} tour ni bar o visit .\\ (Isaac is \textbf{now} visiting the bar for a tour.)\end{tabular} \\\bhline{1.5pt}
\end{tabular}
\caption{Examples of problems that are in the training data (seen problems) and corresponding problems that are not in the training data (unseen problems) in a tense fragment-based split setting. TF means the tense fragment.}
\label{tab:seen_unseen_cat}
\vspace{-5pt}
\end{table*}

\subsubsection{Tense Fragment-Based Split}
Tense fragment refers to the categorization of the problems described in Section \ref{subsec:fragments}.
We define two splits based on the tense fragments: %
\frageasy{} and \fraghard{}.
These splits aim to test whether LMs can learn temporal inference from basic problems and generalize the acquired inference patterns to more challenging problems.
Therefore, both \frageasy{} and \fraghard{} include only basic problems in the training data and challenging problems in the test data.
\fraghard{} contains a higher proportion of challenging problems and fewer tense fragments in the training data, which is a more difficult setting for models.

We define basic and challenging problems based on the sub-tense fragments in the tense fragment classification.
For example, as in the first example in Table~\ref{tab:seen_unseen_cat}, suppose a certain tense fragment has sub-tense fragments that are finer than that tense fragment.
In this case, the original tense fragment (Order relation – Transitive) is considered as basic, and the subcategories (Order relation – Transitive + Before/After) are considered as challenging.
In contrast, as in the second example in Table~\ref{tab:seen_unseen_cat}, if there is no such sub-tense fragment, but there are sub-tense fragments with the same granularity as that of the classification, one (Usage of 現在 (\textit{now}) – Present tense) is considered as basic, and the other (Usage of 現在 (\textit{now}) – Past tense) is considered as challenging.

\begin{table*}[t]
\scriptsize
\centering
\begin{tabular}{l|l|l}\bhline{1.5pt}
       & Seen problem                                                                                                                                                                                                                                                                                                                                                                                                                                                                                                                                            & Unseen problem                                                                                                                                                                                                                                                                                                                                                                                                                                                                                                                                                                                                                                                                                                                                        \\  \hline\hline
 & Format: Year, Gold label: Neutral                                                                                                                                                                                                                                                                                                                                                                                                                                                                                                                                                    & Format: Year-Month-Day-Hour, Gold label: Entailment                                                                                                                                                                                                                                                                                                                                                                                                                                                                                                                                                                                                                                                                                                                                   \\\hline
P      & \begin{tabular}[c]{@{}l@{}}パット\ が\ 6\ 年間\ 以内\ に\ 代価\ を\ 支払った\ 。\\ Pat ga 6 year within ni price o paid .\\ (Pat paid the price within 6 years.)\\ パット\ は\ \textbf{2009\ 年}\ に\ その\ 代価\ を\ 支払い\ 始めた\ 。\\ Pat wa \textbf{2009 year} ni its price o pay began .\\ (Pat began paying the price in \textbf{2009}.)\end{tabular}  & \begin{tabular}[c]{@{}l@{}}エレン\ が\ 2\ 年間\ 以内\ に\ 考え\ を\ 変えた\ 。 \\ Ellen ga 2 years within ni mind o changed .\\ (Ellen changed her mind within 2 years. )\\ エレン\ は\ \textbf{2016\ 年\ 11\ 月\ 18\ 日\ 15\ 時}\ に\ その\ 考え\ を\ 変え\ 始めた\ 。\\ Ellen wa \textbf{2016 year 11 month 18 day 15 hour} ni its mind o change began .\\ (Ellen began to change her mind at \textbf{15:00 on November 18, 2016}.)\end{tabular}                              \\\hline
H      & \begin{tabular}[c]{@{}l@{}}パット\ は\ \textbf{2011\ 年}\ まで\ に\ その\ 代価\ を\ 支払い\ 終えた\ 。\\ Pat wa \textbf{2011 year} until ni its price o pay finished .\\ (Pat finished paying the price by \textbf{2011}.)\end{tabular}                                                                                                                                                                                                                 & \begin{tabular}[c]{@{}l@{}}エレン\ は\ \textbf{2020\ 年\ 10\ 月\ 15\ 日\ 21\ 時}\ まで\ に\ その\ 考え\ を\ 変え\ 終えた\ 。\\ Ellen wa \textbf{2020 year 10 month 15 day 21 hour} until ni its mind wo change finished .\\ (Ellen finished changing her mind by \textbf{21:00 on October 15, 2020}.)\end{tabular}                                                                                                                                                                                                                                                                                                                                                                                                                                                                                                                                                                                                                                                                                                                                                                                                                                                                \\\hline\hline
 & Format: Year-Month, Gold label: Entailment                                                                                                                                                                                                                                                                                                                                                                                                                                                                                                                                             & Format: Year-Month-Day-Hour, Gold label: Entailment                                                                                                                                                                                                                                                                                                                                                                                                                                                                                                                                                                                                                                                                                                                                   \\\hline
P      & \begin{tabular}[c]{@{}l@{}}\textbf{2018\ 年\ 8\ 月}\ 以来\ 、\ ウォルター\ は\ 閣僚\ に\ 指示している\ 。 \\ \textbf{2018 year 8 month} since , Walter wa cabinet ni instruct .\\ (Since \textbf{August 2018}, Walter has instructed cabinet members.)\\ 現在\ 、\ \textbf{2018\ 年\ 11\ 月}\ である\ 。\\ now , \textbf{2018 year 11 month dearu} .\\ (It is now \textbf{November 2018}.)\end{tabular} & \begin{tabular}[c]{@{}l@{}}\textbf{2008\ 年\ 2\ 月\ 27\ 日\ 0\ 時}\ 以来\ 、\ ビクター\ は\ ソフトバンク\ に\ 移籍している\ 。\\ \textbf{2008 year 2 month 27 day 0 hour} since , Victor wa Softbank ni transfer .\\ (Since \textbf{0:00 on February 27, 2008}, Victor has been transferred to Softbank.)\\ 現在\ 、\ \textbf{2008\ 年\ 2\ 月\ 27\ 日\ 4\ 時}\ である\ 。\\ now , \textbf{2008 year 2 month 27 day 4 hour dearu} .\\ (It is now \textbf{4:00 on February 27, 2008}.)\end{tabular} \\\hline
H      & \begin{tabular}[c]{@{}l@{}}ウォルター\ は\ \textbf{2018\ 年\ 9\ 月}\ には\ 閣僚\ に\ 指示していた\ 。\\ Walter wa \textbf{2018 year 9 month} niwa cabinet ni instruct .\\ (Walter had instructed the cabinet ministers in \textbf{September 2018}.)\end{tabular}                                                                                                                                                                                                  & \begin{tabular}[c]{@{}l@{}}ビクター\ は\ \textbf{2008\ 年\ 2\ 月\ 27\ 日\ 1\ 時}\ には\ ソフトバンク\ に\ 移籍していた\ 。\\ Victor wa \textbf{2008 year 2 month 27 day 1 hour} niwa Softbank ni transfer .\\ (Victor was transferred to Softbank at \textbf{1:00 on February 27, 2008}.)\end{tabular}                                                                                                                                                                                                                                                                                                                                                                                                                                                                                                                                                                                                                                                                                                                                              \\ \bhline{1.5pt}
\end{tabular}
\caption{Examples of problems that are in the training data (seen problems) and corresponding problems that are not in the training data (unseen problems) in a time format-based split setting.}
\label{tab:seen_unseen_tf}
\vspace{-10pt}
\end{table*}

\subsubsection{Time Format-Based Split}
\label{subsubsec:timeformat}
Time format represents the format of the temporal expression inserted in a problem.
In this study, we define ten time formats by combining multiple time units (year, month, day, and hour) for time points and define two splits based on the time formats.
This split aims to test whether LMs can learn the size relationships between time units (year > month > day > hour) from a minimal number of combinations of units and generalize the acquired inference patterns to apply them to complex combinations.

The first split is \formhard{}, which contains only a single time unit pattern (i.e., patterns involving only year, only month, only day, or only hour) in a training set and evaluates models on combined patterns of multiple time units.

The other split is \formeasy{}, which includes a minimum number of combinations (i.e., year-month pattern, month-day pattern, and day-hour pattern) that allow the models to understand the size relationships between time units, as shown in the second example in Table~\ref{tab:seen_unseen_tf}.
By comparing the accuracy of \formeasy{} and \formhard{}, we can determine whether LMs can learn and generalize the size relationships between time units.

\begin{table*}[t]
\scriptsize
\centering
\scalebox{0.97}{
\begin{tabular}{l|l|l}\bhline{1.5pt}
       & Seen problem                                                                                                                                                                                                                                                                                                                                                                                                                                                                                                                                            & Unseen problem                                                                                                                                                                                                                                                                                                                                                                                                                                                                                                                                                                                                                                                                                                                                        \\ \hline\hline
 & Span: Random, Gold label: Neutral                                                                                                                                                                                                                                                                                                                                                                                                                                                                                                                                             & Span: Short, Gold label: Contradiction                                                                                                                                                                                                                                                                                                                                                                                                                                                                                                                                                                                                                                                                                                                                   \\\hline
P      & \begin{tabular}[c]{@{}l@{}}\textbf{2002\ 年\ 8\ 月\ 16\ 日\ 7\ 時}\ 以来\ 、\ ウォルター\ は\ 実家\ に\ 泊まっている\ 。 \\ \textbf{2018 year 8 month 16 day 7 hour} since , Walter wa parents' house ni stay .\\ (Walter has been staying at his parents' house since \textbf{7:00 on August 16, 2002}.)\\ 現在\ 、\ \textbf{2013\ 年\ 5\ 月\ 26\ 日\ 3\ 時}\ である\ 。\\ now , \textbf{2013 year 5 month 26 day 3 hour} dearu .\\ (It is now \textbf{3:00 on May 26, 2013}.)\end{tabular} & \begin{tabular}[c]{@{}l@{}}\textbf{2015\ 年\ 9\ 月\ 11\ 日\ 7\ 時}\ 以来\ 、\ フランク\ は\ 細工\ に\ 挑戦している\ 。\\ \textbf{2015 year 9 month 11 day 7 hour} since , Frank wa craft ni try .\\ (Frank has been trying to craft since \textbf{7:00 on September 11, 2015}.)\\ 現在 、 \textbf{2015 年 9 月 11 日 10 時} である 。\\ now , \textbf{2015 year 9 month 11 day 10 hour} dearu .\\ (It is now \textbf{10:00 on September 11, 2015}.)\end{tabular} \\\hline
H      & \begin{tabular}[c]{@{}l@{}}ウォルター\ は\ \textbf{2018\ 年\ 5\ 月\ 15\ 日\ 12\ 時}\ には\ 実家\ に\ 泊まっていた\ 。\\ Walter wa \textbf{2018 year 5 month 15 day 12 hour} niwa parents' house ni stay .\\ (Walter was staying at his parents' house at \textbf{12:00 on May 15, 2018}.)\end{tabular}                                                                                                                                                                                                  & \begin{tabular}[c]{@{}l@{}}フランク\ は\ \textbf{2015\ 年\ 9\ 月\ 11\ 日\ 5\ 時}\ には\ 細工\ に\ 挑戦していた\ 。\\ Frank wa \textbf{2015 year 9 month 11 day 5 hour} niwa craft ni try .\\ (Frank was trying to craft at \textbf{5:00 on September 11, 2015}.)\end{tabular}                                                                                                                                                                                                                                                                                                                                                                                                                                                                                                                                                                                                                                                                                                                                              \\\bhline{1.5pt}
\end{tabular}
}
\vspace{-5pt}
\caption{Examples of problems that are in the training data (seen problems) and corresponding problems that are not in the training data (unseen problems) in a time span-based split setting.}
\label{tab:seen_unseen_ts}
\vspace{-10pt}
\end{table*}

\subsubsection{Time Span-Based Split}
Time span represents the closeness of temporal expressions when multiple temporal expressions appear in a problem.
In this study, we define two time spans: \short{} and \random{}.
In \short{} time span problems, the temporal expressions are generated such that the time points included in the problem are close to each other (see Appendix~\ref{append:temporal_expression_generation}), as shown in the unseen problem in Table~\ref{tab:seen_unseen_ts}.
On the other hand, in \random{} time span problems, the distance between the time points included in the problem is not predetermined, and the temporal expressions are generated in the same manner as described in Section~\ref{subsec:problemgeneration}.
Therefore, the distances between the time points included in a problem are often far apart, as shown in the seen problem in Table~\ref{tab:seen_unseen_ts}.

When a model determines the order of two time points, the model must compare the two time points in order, starting with the largest unit.
If two time points are far apart, then the model can determine their order by comparing only the larger units, but if two time points are close, then the model must compare additional units to determine their order.
For example, the order of January 1, 2010, at 1:00 and October 10, 2020, at 10:00 can be determined by looking only at the year, but the order of January 1, 2010, at 1:00 and January 1, 2010, at 10:00 requires comparing the year, month, day, and hour in order.
Therefore, we consider that determining the order relationships between close time points is more difficult than determining the order relationships between distant time points.

We define a time span-based split that contains only \random{} in the training data.
This split aims to test whether LMs can learn the order relationships of temporal expressions and generalize the acquired inference patterns to apply them to combinations of temporal expressions that require more difficult evaluation.

\begin{table*}[t]
\centering
\tiny
\begin{tabular}{ccc|c|c|c|cc|ccc|c}
\bhline{1.5pt}
\multicolumn{2}{c|}{\multirow{3}{*}{Model}} &
\multicolumn{1}{c|}{\multirow{3}{*}{\begin{tabular}[c]{@{}c@{}}seen/\\ unseen\end{tabular}}} &
\multicolumn{2}{c|}{Zero-shot} &
\multicolumn{6}{c}{Fine-tuning}
\\ \cline{4-12} 
\multicolumn{2}{c|}{} &
\multicolumn{1}{c|}{} &
\multirow{2}{*}{\begin{tabular}[c]{@{}c@{}}Mono\\ lingual\end{tabular}} &
\multirow{2}{*}{\begin{tabular}[c]{@{}c@{}}Cross-\\ lingual\end{tabular}} &
\multicolumn{1}{c|}{\multirow{2}{*}{\begin{tabular}[c]{@{}c@{}}IID\\ Split\end{tabular}}} &
\multicolumn{2}{c|}{Tense Fragment} &
\multicolumn{3}{c|}{Time Format} &
\multicolumn{1}{c}{\multirow{2}{*}{\begin{tabular}[c]{@{}c@{}}Time\\ Span\end{tabular}}}
\\ \cline{7-11} 
\multicolumn{2}{c|}{} &
\multicolumn{1}{c|}{} & & & 
\multicolumn{1}{c|}{} &
Easy &
\multicolumn{1}{c|}{Hard} &
Easy &
Hard &
\multicolumn{1}{c|}{$\Delta$} &
\multicolumn{1}{c}{} 
\\ \hline
\multicolumn{1}{c|}{\multirow{6}{*}{BERT}} &
\multicolumn{1}{c|}{\multirow{3}{*}{base}}  &
seen &
- &
- &
.891$_{\pm 0.02}$ &
.879$_{\pm 0.01}$ & 
.812$_{\pm 0.05}$ &
.839$_{\pm 0.02}$ & 
.800$_{\pm 0.02}$ & 
.039$_{\pm 0.03}$ &
.757$_{\pm 0.03}$
\\ \cline{3-3}
\multicolumn{1}{c|}{} &
\multicolumn{1}{c|}{}  &
unseen &
.428$_{\pm 0.02}$ &
- &
- &
.405$_{\pm 0.04}$ &
.379$_{\pm 0.02}$ &
.897$_{\pm 0.03}$ &
.761$_{\pm 0.04}$ &
\textbf{.136$_{\pm 0.05}$} & 
.662$_{\pm 0.05}$
\\ \cline{3-3}
\multicolumn{1}{c|}{} &
\multicolumn{1}{c|}{}  &
$\Delta$ &
- &
- &
- &
\textbf{.474$_{\pm 0.04}$} &
\textbf{.433$_{\pm 0.05}$} &
- &
- &
- &
\textbf{.095$_{\pm 0.06}$}
\\ \cline{2-12}
\multicolumn{1}{c|}{} &
\multicolumn{1}{c|}{\multirow{3}{*}{large}} &
seen &
- &
- &
.955$_{\pm 0.01}$ &
.969$_{\pm 0.01}$ &
.968$_{\pm 0.02}$ &
.920$_{\pm 0.02}$ &
.922$_{\pm 0.01}$ &
-.002$_{\pm 0.02}$ &
.912$_{\pm 0.01}$
\\ \cline{3-3}
\multicolumn{1}{c|}{} &
\multicolumn{1}{c|}{}  &
unseen &
.440$_{\pm 0.03}$ &
- &
- &
.457$_{\pm 0.03}$ &
.419$_{\pm 0.01}$ &
.970$_{\pm 0.02}$ &
.893$_{\pm 0.02}$ &
\textbf{.077$_{\pm 0.03}$} &
.876$_{\pm 0.04}$
\\ \cline{3-3}
\multicolumn{1}{c|}{} &
\multicolumn{1}{c|}{}  &
$\Delta$ &
- &
- &
- &
\textbf{.512$_{\pm 0.03}$} &
\textbf{.549$_{\pm 0.02}$} &
- &
- &
- &
\textbf{.036$_{\pm 0.04}$}
\\ \hline
\multicolumn{1}{c|}{\multirow{6}{*}{RoBERTa}} &
\multicolumn{1}{c|}{\multirow{3}{*}{base}}  &
seen &
- &
- &
.914$_{\pm 0.02}$ &
.898$_{\pm 0.03}$ &
.851$_{\pm 0.07}$ &
.832$_{\pm 0.03}$ &
.754$_{\pm 0.08}$ &
.078$_{\pm 0.09}$ &
.749$_{\pm 0.06}$
\\ \cline{3-3}
\multicolumn{1}{c|}{} &
\multicolumn{1}{c|}{}  &
unseen &
.468$_{\pm 0.03}$ &
- &
- &
.388$_{\pm 0.02}$ &
.318$_{\pm 0.02}$ &
.846$_{\pm 0.04}$ &
.677$_{\pm 0.12}$ &
\textbf{.169$_{\pm 0.13}$} &
.669$_{\pm 0.05}$
\\ \cline{3-3}
\multicolumn{1}{c|}{} &
\multicolumn{1}{c|}{}  &
$\Delta$ &
- &
- &
- &
\textbf{.510$_{\pm 0.04}$} &
\textbf{.533$_{\pm 0.07}$} &
- &
- &
- &
\textbf{.080$_{\pm 0.08}$}
\\ \cline{2-12}
\multicolumn{1}{c|}{} &
\multicolumn{1}{c|}{\multirow{3}{*}{large}} &
seen &
- &
- &
.937$_{\pm 0.03}$ &
.970$_{\pm 0.01}$ &
.984$_{\pm 0.01}$ &
.914$_{\pm 0.03}$ &
.907$_{\pm 0.01}$ &
.007$_{\pm 0.03}$ &
.819$_{\pm 0.13}$
\\ \cline{3-3}
\multicolumn{1}{c|}{} &
\multicolumn{1}{c|}{}  &
unseen &
.460$_{\pm 0.02}$ &
- &
- &
.445$_{\pm 0.03}$ &
.399$_{\pm 0.04}$ &
.967$_{\pm 0.02}$ &
.884$_{\pm 0.01}$ &
\textbf{.083$_{\pm 0.02}$} &
.799$_{\pm 0.11}$
\\ \cline{3-3}
\multicolumn{1}{c|}{} &
\multicolumn{1}{c|}{}  &
$\Delta$ &
- &
- &
- &
\textbf{.525$_{\pm 0.03}$} &
\textbf{.585$_{\pm 0.04}$} &
- &
- &
- &
\textbf{.020$_{\pm 0.17}$}
\\ \hline
\multicolumn{1}{c|}{\multirow{6}{*}{\begin{tabular}[c]{@{}c@{}}XLM-\\ RoBERTa\end{tabular}}} &
\multicolumn{1}{c|}{\multirow{3}{*}{base}}  &
seen &
- &
- &
.768$_{\pm 0.05}$ &
.683$_{\pm 0.01}$ &
.649$_{\pm 0.02}$ &
.690$_{\pm 0.09}$ &
.607$_{\pm 0.02}$ &
.083$_{\pm 0.09}$ &
.553$_{\pm 0.06}$
\\ \cline{3-3}
\multicolumn{1}{c|}{} &
\multicolumn{1}{c|}{}  &
unseen &
- &
.411$_{\pm 0.03}$ &
- &
.238$_{\pm 0.01}$ &
.309$_{\pm 0.02}$ &
.678$_{\pm 0.06}$ &
.541$_{\pm 0.01}$ &
\textbf{.137$_{\pm 0.06}$} &
.553$_{\pm 0.06}$
\\ \cline{3-3}
\multicolumn{1}{c|}{} &
\multicolumn{1}{c|}{}  &
$\Delta$ &
- &
- &
- &
\textbf{.445$_{\pm 0.01}$} &
\textbf{.340$_{\pm 0.03}$} &
- &
- &
- &
\textbf{.000$_{\pm 0.08}$}
\\ \cline{2-12}
\multicolumn{1}{c|}{} &
\multicolumn{1}{c|}{\multirow{3}{*}{large}} &
seen &
- &
- &
.941$_{\pm 0.01}$ &
.952$_{\pm 0.02}$ &
.955$_{\pm 0.03}$ &
.883$_{\pm 0.05}$ &
.862$_{\pm 0.06}$ &
.021$_{\pm 0.08}$ &
.761$_{\pm 0.08}$
\\ \cline{3-3}
\multicolumn{1}{c|}{} &
\multicolumn{1}{c|}{}  &
unseen &
- &
.488$_{\pm 0.03}$ &
- &
.455$_{\pm 0.04}$ &
.383$_{\pm 0.02}$ &
.935$_{\pm 0.06}$ &
.783$_{\pm 0.08}$ &
\textbf{.152$_{\pm 0.10}$} &
.735$_{\pm 0.09}$
\\ \cline{3-3}
\multicolumn{1}{c|}{} &
\multicolumn{1}{c|}{}  &
$\Delta$ &
- &
- &
- &
\textbf{.497$_{\pm 0.04}$} &
\textbf{.572$_{\pm 0.04}$} &
- &
- &
- &
\textbf{.026$_{\pm 0.12}$}
\\ \bhline{1.5pt}
\end{tabular}
\caption{Results on our test data (average accuracy and standard deviation of five runs).}
\label{tab:main_results}
\vspace{-15pt}
\end{table*}
\section{Experiments}
\label{sec:experiments}
We evaluate several NLI models on our dataset.
We consider six pre-trained LMs (Japanese BERT-base/large, Japanese RoBERTa-base/large, multilingual XLM-RoBERTa-base/large)\footnote{We did not evaluate the prompt-tuning models such as GPT-3 because accurate comparisons with other models in the fine-tuning setting are difficult.} available on huggingface/transformers\footnote{\url{https://huggingface.co/transformers/}} in our experiments.
We conduct experiments in three settings: zero-shot (monolingual), zero-shot (cross-lingual), and fine-tuning.
Here, zero-shot means that we do not use our training data but use existing Japanese NLI datasets for training data.
The statistics of the datasets used in our experiments are provided in Appendix~\ref{appendix:data_statistics}.
\paragraph{Zero-shot setting (monolingual)}
We train the LMs on three concatenated NLI datasets: the standard Japanese NLI datasets JSNLI (automatic translation of the English SNLI dataset~\citep{bowman-etal-2015-large})~\citep{yoshikoshi-etal-2020-multilingualization} and JSICK (manual translation of the English SICK dataset~\citep{marelli-etal-2014-sick})~\citep{yanaka-mineshima-2022-compositional}, and the Japanese NLI dataset PLMUTE\_ja~\citep{sugimoto-yanaka-2022-compositional}, which involves temporal order.
We then evaluate the models on our test data.
\paragraph{Zero-shot setting (cross-lingual)}
We train the LMs on three concatenated NLI datasets: the standard English NLI dataset SNLI, SICK, and the English NLI dataset PLMUTE~\citep{thukral-etal-2021-probing}, which involves temporal order and duration.
We then evaluate the models on our test data.
\paragraph{Fine-tuning setting}
We train and evaluate the LMs on our training data and test data.

Additionally, in the fine-tuning setting, we train the LMs on the split training data described in Section \ref{subsec:splitproblem}, as well as on all of the training data.

In all experiments, we conduct five trials and calculate the averages and standard deviations of the accuracy of the models.
Training details are provided in Appendix~\ref{appendix:training_details}.

\begin{table*}[t]
\scriptsize
\centering
\begin{tabular}{l|l|l}\bhline{1.5pt}
 &
Seen problem &
Unseen problem\\\hline\hline
 &
\begin{tabular}[c]{@{}l@{}}
TF: Habituality - Unmentioned TP + Always\\
Gold label: Neutral
\end{tabular} &
\begin{tabular}[c]{@{}l@{}}
TF: Habituality + Negation - Unmentioned TP + Always\\
Gold label: Contradiction, Pred label: Neutral
\end{tabular}\\\hline
P &
\begin{tabular}[c]{@{}l@{}}
イヴァン\ は\ いつも\ 図面\ を\ 遅れて\ 出す\ 。 \\
Ivan wa always drawing o late submit .\\
(Ivan always submits his drawing late. )\\
2011\ 年\ 11\ 月\ 28\ 日\ 16\ 時\ に\ イヴァン\ は\ 図面\ を\ 出した\ 。\\
2011 year 11 month 28 day 16 hour ni Ivan wa drawing o submit .\\ 
(Ivan submitted his drawing at 16:00 on November 28, 2011.)
\end{tabular} &
\begin{tabular}[c]{@{}l@{}}
デイヴ\ は\ いつも\ マンション\ を\ 遅れて\ 訪れる\ 。\\
Dave wa always apartment o late visit .\\
(Dave always visits the apartment late.)\\
2002\ 年\ 5\ 月\ 11\ 日\ 14\ 時\ に\ デイヴ\ は\ マンション\ を\ 訪れた\ 。\\
2002 year 5 month 11 day 14 hour ni Dave wa apartment o visit .\\
(Dave visited the apartment on May 11, 2002 at 14:00.)
\end{tabular}\\\hline
H &
\begin{tabular}[c]{@{}l@{}}
イヴァン\ は\ 2011\ 年\ 11\ 月\ 28\ 日\ 22\ 時\ に\ 図面\ を\ 遅れて\ 出した\ 。\\
Ivan wa 2011 year 11 month 28 day 22 hour ni drawing o late submit .\\
(Ivan submitted his drawing late at 22:00 on November 28, 2011.)
\end{tabular} &
\begin{tabular}[c]{@{}l@{}}
デイヴ\ は\ 2012\ 年\ 2\ 月\ 1\ 日\ 0\ 時\ に\ マンション\ を\ \textbf{遅れ\ ず}\ に\ 訪れた\ 。\\
Dave wa 2012 year 2 month 1 day 0 hour ni apartment o late \textbf{not} ni visit .\\
 (Dave visited the apartment on February 1, 2012 at 0:00 \textbf{without} delay.)
\end{tabular}\\
\bhline{1.5pt}
\end{tabular}
\caption{An example of unseen problem that RoBERTa-large could not solve in \frageasy{} and the corresponding seen problem in the training data. TF means the tense fragment.}
\label{tab:error_example}
\vspace{-15pt}
\end{table*}

\section{Results and Discussion}
\label{sec:results_and_discussion}
Table \ref{tab:main_results} shows the results of all our experiments.
Overall, monolingual models with larger model sizes tend to perform better.
In this section, we describe the results for each setting in detail.

\subsection{Zero-shot setting}
The two left columns in Table \ref{tab:main_results} show the results on the zero-shot setting.
As Table \ref{tab:main_results} shows, the accuracy of both the monolingual and cross-lingual models is approximately 40\%, and there is no significant difference between them.
One possible reason is that SNLI, SICK, and their Japanese versions (JSNLI and JSICK) do not contain temporal inference, and the temporal inference patterns obtained from PLMUTE are only a fraction of the inference patterns required to solve our test set.

\subsection{Fine-tuning setting}
The right side of Table \ref{tab:main_results} shows the results on the fine-tuning setting.
As expected, all models are highly accurate on the IID split setting (i.e., the setting in which all training data were used).
We then discuss the results of the experiments using the splits described in Section \ref{subsec:splitproblem}.

\paragraph{Tense Fragment-based Split}
In the tense fragment-based split, the difference in accuracy between seen and unseen problems was nearly 50\% for all models on both \frageasy{} and \fraghard{}.
This suggests that the models cannot generalize the temporal inferences obtained from the training data.

Table~\ref{tab:error_example} shows an example of unseen problems that RoBERTa-large could not solve on \frageasy{} and the corresponding seen problems in the training data.
Because all models obtained similar results in relation to the generalization ability of LMs for temporal inference, we focus on the RoBERTa-large model, which achieved the best performance on our dataset.
For this example, the model gave the same prediction for the both unseen and seen problems.
The other tense fragment problems that the model could not solve on \frageasy{} have the same characteristics.
Specifically, the model tended to predict incorrect labels for problems in which the premises and hypotheses of seen and unseen problems were very similar (differences are highlighted in bold), but the gold labels were different, as shown in Table~\ref{tab:error_example}.
This suggests that this model does not capture the essential meaning of a sentence but determines the entailment relations based only on superficial information (i.e., the model does not generalize temporal inference patterns).

\paragraph{Time Format-based Split}
As shown in Table \ref{tab:main_results} shows, all models except XLM-RoBERTa-base achieved 80\% accuracies on both unseen problems and seen problems of \formeasy{}.
Furthermore, detailed analysis revealed that the XLM-RoBERTa-base did not solve problems that required inference of the size relationships between time units.
This indicates that XLM-RoBERTa-base only fails to generalize the size relation between time units.
One potential reason for this is that this model is cross-lingual and not large.
In contrast, on \formhard{}, all models exhibited reduced accuracy for the unseen problems compared to the seen problems.
This indicates that the models do not have a priori knowledge regarding the size relationships between time units.
Therefore, we consider that on \formeasy{}, BERT and RoBERTa succeeded in generalizing the inference patterns of the size relationships between time units based on minimal combinations of time units in the training data.

\paragraph{Time Span-based Split}
On the time span-based split, the large models achieved comparable accuracy on both the seen and unseen problems, whereas the base models tended to exhibit lower accuracy on the unseen problems.
This suggests that the large models can generalize methods for determining the order relationships between time points, but the base models cannot generalize.

\section{Conclusion}
\label{sec:conclusion}
In this study, we constructed \oursno, a temporal Japanese NLI dataset, using a template-based approach.
Our dataset is controllable in terms of difficulty, vocabulary, and size based on this approach.
We conducted experiments using our dataset to probe the generalization ability of pre-trained language models for temporal inference.
The experimental results indicated that current LMs can generalize for time format splits and time span splits but fail to generalize for tense fragment splits.
Our dataset demonstrates that there is room for improvement in the generalization ability of current standard LMs for temporal inference.
Because our method is applicable to the construction of datasets for other linguistic phenomena (e.g., modality, comparative), we plan to investigate the generalization ability of language models for other phenomena using the template-based approach in the future.

\section{Limitations}
\label{sec:limitations}
In this section, we discuss two limitations of this study.
The first limitation is that aspect and temporal commonsense are outside the scope of our dataset.
Here, temporal commonsense refers to knowledge regarding events and the appropriate duration of those events.
For example, the event ``I washed my face for three years'' is unnatural in terms of temporal commonsense, but this study did not consider such unnaturalness.

The second limitation is that the proposed method is currently applicable only to Japanese.
In this study, we used a Japanese case frame dictionary to generate natural sentences.
However, other languages such as English do not have resources equivalent to such a dictionary.
Therefore, to apply our method to additional languages, we must first prepare a case frame dictionary for each language.

\section*{Acknowledgements}
We thank the two anonymous reviewers for their helpful comments and suggestions, which improved this paper.
This work was supported by JST, PRESTO grant number JPMJPR21C8, Japan.

\bibliography{anthology,custom}
\bibliographystyle{acl_natbib}

\newpage
\appendix
\section{Tense Fragment}
\label{append:tense_fragment}

Table \ref{tab:category} shows the tense fragments we defined.

\begin{table}[h]
\small
\centering
\begin{tabular}{l|l}
\bhline{1.5pt}
Tense Fragment                           & Sub-tense Fragment                  \\ \hline\hline
Temporal commonsense               & Usage of 現在 (\textit{now})      \\ \hline
\multirow{2}{*}{Temporal ordering} & Continuity of state          \\ \cline{2-2} 
                                   & Ordering relation            \\ \hline
\multirow{2}{*}{Time point}           & Mentioned time point            \\ \cline{2-2} 
                                   & Unmentioned time point          \\ \hline
Temporal anaphora                  & \begin{tabular}[c]{@{}l@{}}Reference resolution\\ of 昨日 (\textit{yesterday})\end{tabular} \\ \hline
\multirow{2}{*}{Interval}          & Comparison of two intervals  \\ \cline{2-2} 
                                   & Completion of eventuality    \\ \hline
\multirow{4}{*}{Habituality}       & Mentioned time point            \\ \cline{2-2} 
                                   & Unmentioned time point          \\ \cline{2-2} 
                                   & Negation                     \\ \cline{2-2} 
                                   & Existential quantification   \\ \bhline{1.5pt}
\end{tabular}
\caption{Tense fragments we introduced in this study.}
\label{tab:category}
\vspace{-10pt}
\end{table}

\section{Problem Creation for Some JSeM Problems}
\begin{table*}[t]
\scriptsize
\centering
\begin{tabular}{l|l|l}\bhline{1.5pt}
&
Original problem &
New problem \\\hline\hline
&
Gold label: Entailment &
Gold label: Contradiction \\\hline
P &
\begin{tabular}[c]{@{}l@{}}
スミス\ は\ ジョーンズ\ が\ 去る\ 前\ に\ 去った\ 。\\
Smith wa Jones ga leave before ni leave .\\
(Smith left before Jones left.)\\
ジョーンズ\ は\ アンダーソン\ が\ 去る\ 前\ に\ 去った\ 。\\
Jones wa Anderson ga leave before ni leave .\\
(Jones left before Anderson left.)
\end{tabular} &
\begin{tabular}[c]{@{}l@{}}
スミス\ は\ ジョーンズ\ が\ 去る\ 前\ に\ 去った\ 。\\
Smith wa Jones ga leave before ni leave .\\
(Smith left before Jones left.)\\
ジョーンズ\ は\ アンダーソン\ が\ 去る\ 前\ に\ 去った\ 。\\
Jones wa Anderson ga leave before ni leave .\\
(Jones left before Anderson left.)
\end{tabular}\\\hline
H &
\begin{tabular}[c]{@{}l@{}}
スミス\ は\ アンダーソン\ が\ \textbf{去る\ 前}\ に\ 去った\ 。\\
Smith wa Anderson ga leave \textbf{before} ni leave .\\
(Smith left \textbf{before} Anderson left.)
\end{tabular} &
\begin{tabular}[c]{@{}l@{}}
スミス\ は\ アンダーソン\ が\ \textbf{去った\ 後}\ に\ 去った\ 。\\
Smith wa Anderson ga leave \textbf{after} ni leave .\\
(Smith left \textbf{after} Anderson left.)
\end{tabular} \\\hline\hline
&
Gold label: Neutral &
Gold label: Entailment \\\hline
P &
\begin{tabular}[c]{@{}l@{}}
スミス\ が\ 2\ 時間\ \textbf{以内\ に}\ 報告書\ を\ 書いた\ 。\\
Smith ga 2 hour \textbf{within} ni report o write .\\
(Smith wrote a report \textbf{within} two hours.)
\end{tabular}
&
\begin{tabular}[c]{@{}l@{}}
スミス\ が\ 2\ 時間\ \textbf{で}\ 報告書\ を\ 書いた\ 。\\
Smith ga 2 hour \textbf{de} report o write .\\
(Smith wrote a report \textbf{in} two hours.)
\end{tabular} \\\hline
H &
\begin{tabular}[c]{@{}l@{}}
スミス\ は\ その\ 報告書\ を\ 書く\ の\ に\ 2\ 時間\ を\ 費やした\ 。\\
Smith wa that report o write no ni 2 hour o spent .\\
(Smith spent two hours writing that report.)
\end{tabular}
&
\begin{tabular}[c]{@{}l@{}}
スミス\ は\ その\ 報告書\ を\ 書く\ の\ に\ 2\ 時間\ を\ 費やした\ 。\\
Smith wa that report o write no ni 2 hour o spent .\\
(Smith spent two hours writing that report.)
\end{tabular} \\\bhline{1.5pt}
\end{tabular}
\caption{Examples of created problems and corresponding original problems in JSeM.}
\label{tab:new_problem}
\vspace{-10pt}
\end{table*}
\label{appendix:new_template}
Table~\ref{tab:new_problem} shows examples of created problems and corresponding original problems in JSeM.
As shown in Table~\ref{tab:new_problem}, original and new problems are similar but have different gold labels.
We also create templates for these created problems.

\section{Temporal Expression Generation in \short{} Time Span}
\label{append:temporal_expression_generation}

The temporal expressions in \short{} are generated as follows.
In the case of generating intervals, they are generated as described in Section~\ref{subsec:problemgeneration}, except that the integer selection range is one to three instead of one to nine.
In the case of generating time points, we first identify the next largest unit after the smallest unit of the time format in the current problem and then calculate the duration of one-third of that unit.
We then determine a selection range from a randomly selected time point to a time point that is advanced by the calculated duration.
For example, if the smallest unit is ``hour,'' then the next smallest unit is ``day,'' so the selection range is between a specific time point and another time point one-third of a day (eight hours) in the future.

\section{Details for Dataset Artifacts Analysis}
\label{appendix:detail_artifact}
As mentioned in Section~\ref{subsubsec:dataset_artifacts}, dataset artifacts analysis reveals correlations between labels and specific words.
Formally, this analysis is a one-side binomial hypothesis test with the null hypothesis $p(y|x_i) = 1/3$, where $y \in \{\textit{Entailment}, \textit{Neutral}, \textit{Contradiction}\}$, and $x_i$ is a word included in the vocabulary.
For this analysis, we first split the hypothesis and premise sentences into individual words/tokens using Juman++~\citep{morita-etal-2015-morphological}.
We then count the number of occurrences of the gold label $y$ in the $n_i$ examples for every word $x_i$ present in those examples.
$p(y|x_i)$ is estimated based on the fraction of the count of the gold label $y$ over $n_i$.
According to the protocol described in \citet{gardner-etal-2021-competency}, the null hypothesis is either accepted or rejected with a significance level of $\alpha = 0.01$ based on the Bonferroni correction.

\section{Data Statistics}
\label{appendix:data_statistics}
Table~\ref{tab:jamp_statistics} shows \ours dataset statistics.
\begin{table}[t]
\centering
\small
\begin{tabular}{lc}\bhline{1.5pt}
Section &
Size \\\hline\hline
\begin{tabular}[l]{@{}l@{}}Train\end{tabular} & 
\begin{tabular}[c]{@{}c@{}}9,750\\ (3,050/3,340/3,360)\end{tabular} \\\hline
\begin{tabular}[l]{@{}l@{}}Test\end{tabular} &
\begin{tabular}[c]{@{}c@{}}344\\ (114/112/118)\end{tabular}
\\ \bhline{1.5pt}
\end{tabular}
\caption{\ours dataset statistics. The lower row in parentheses shows the number of entailment, contradiction, and neutral examples, respectively.}
\label{tab:jamp_statistics}
\end{table}
Table~\ref{tab:other_dataset_statistics} shows sizes of datasets used in our experiments.
\begin{table}[t]
\centering
\small
\begin{tabular}{ll}\bhline{1.5pt}
Dataset Name & Size\\\hline\hline
SNLI~\citep{bowman-etal-2015-large}         & 550,152         \\
SICK~\citep{marelli-etal-2014-sick}         & 9,840           \\
PLMUTE~\citep{thukral-etal-2021-probing}       & 72,720          \\
JSNLI~\citep{yoshikoshi-etal-2020-multilingualization}        & 533,005         \\
JSICK~\citep{yanaka-mineshima-2022-compositional}        & 5,000           \\
PLMUTE\_ja~\citep{sugimoto-yanaka-2022-compositional}   & 11,220          \\\bhline{1.5pt}
\end{tabular}
\caption{Statistics of dataset used in our experiments}
\label{tab:other_dataset_statistics}
\end{table}

\section{Training Details}
\label{appendix:training_details}
We select the best learning rate among [6e-6,8e-6,1e-5,1.2e-5,2e-5] based on the development set.
We use a batch size of 16 for training and eight for test.

\section{Data Licensing}
Japanese case frame dictionary is distributed by Gengo-Shigen-Kyokai.
JSeM is licensed under by BSD-3-Clause license.
Our use of these two datasets is consistent with the terms of the license.

\end{document}